%% file: main.tex
\documentclass[10pt,twocolumn,letterpaper]{article}

\usepackage[pagenumbers]{cvpr} % To force page numbers, e.g. for an arXiv version

\usepackage{graphicx}
\usepackage{amsmath}
\usepackage{amssymb}
\usepackage{booktabs}
\usepackage{xcolor}
\usepackage{multirow}
\usepackage{balance}
\usepackage{soul}
\usepackage{tabulary}
\usepackage{enumitem}
\usepackage[T1]{fontenc}
\usepackage[utf8]{inputenc}
\newcolumntype{x}[1]{>{\centering\arraybackslash}p{#1pt}}
\newlength\savewidth

\usepackage[pagebackref,breaklinks,colorlinks]{hyperref}

\newif\ifshowcomments
\newcommand{\done}[1]{}
\newcommand{\gunnardone}[1]{}
\newcommand{\jessedone}[1]{}
\newcommand{\gauravdone}[1]{}
\newcommand{\robinsondone}[1]{}

\showcommentstrue  % comment this out to hide comments
\ifshowcomments
    \newcommand{\gunnar}[1]{{\color{violet}[gunnar: #1]}}
    \newcommand{\jesse}[1]{{\color{orange}[jesse: #1]}}
    \newcommand{\gaurav}[1]{{\color{purple}[gaurav: #1]}}    
    \newcommand{\robinson}[1]{{\color{teal}[robinson: #1]}}
    
\else
    \newcommand{\gunnar}[1]{}
    \newcommand{\jesse}[1]{}
    \newcommand{\gaurav}[1]{}    
    \newcommand{\robinson}[1]{}

\fi

\newcommand{\ourmodel}{RREx-BoT}
\newcommand{\ourmodelfull}{Remote Referring Expressions with a Bag of Tricks}
\newcommand{\osman}{OSMaN}
\newcommand{\textquote}[1]{\textit{#1}}

\usepackage{colortbl}
\usepackage{xcolor}
\definecolor{LightBlue}{rgb}{0.9, 1, 1}
\newcolumntype{i}{>{\columncolor{LightBlue}}r}
\definecolor{LightRed}{rgb}{1, 0.9, 0.9}
\newcolumntype{j}{>{\columncolor{LightRed}}r}
\definecolor{Gray}{gray}{0.9}
\newcolumntype{k}{>{\columncolor{Gray}}c}
\definecolor{Gray}{gray}{0.9}
\newcolumntype{o}{>{\columncolor{Gray}}r}

\usepackage[capitalize]{cleveref}
\crefname{section}{Sec.}{Secs.}
\Crefname{section}{Section}{Sections}
\Crefname{table}{Table}{Tables}
\crefname{table}{Tab.}{Tabs.}

\def\cvprPaperID{8343} % *** Enter the CVPR Paper ID here
\def\confName{CVPR}
\def\confYear{2023}

\begin{document}

%%%%%%%%% TITLE
\title{RREx-BoT: Remote Referring Expressions with a Bag of Tricks}

\newcommand{\myurl}[1]{\url{#1}}
\author{Gunnar A. Sigurdsson \ \ \ \ 
Jesse Thomason \ \ \ 
Gaurav S. Sukhatme \ \ \ 
Robinson Piramuthu\vspace{0.2cm}\\ 
Amazon Alexa AI
%\\ \myurl{github.com/}\vspace{-0.5cm}
}
\maketitle

%%%%%%%%% ABSTRACT
\begin{abstract}
Household robots operate in the same space for years. 
Such robots incrementally build dynamic maps that can be used for tasks requiring remote object localization.
However, benchmarks in robot learning often test generalization through inference on tasks in unobserved environments.
In an observed environment, locating an object is reduced to choosing from among all object proposals in the environment, which may number in the 100,000s.
Armed with this intuition, using only a generic vision-language scoring model with minor modifications for 3d encoding and operating in an embodied environment, we demonstrate an absolute performance gain of 9.84\% on remote object grounding above state of the art models for REVERIE and of 5.04\% on FAO.
When allowed to pre-explore an environment, we also exceed the previous state of the art pre-exploration method on REVERIE. 
Additionally, we demonstrate our model on a real-world TurtleBot platform, highlighting the simplicity and usefulness of the approach.
Our analysis outlines a ``bag of tricks'' essential for accomplishing this task, from utilizing 3d coordinates and context, to generalizing vision-language models to large 3d search spaces.
\end{abstract}

%%%%%%%%% BODY TEXT
\section{Introduction}
\label{sec:intro}
\input{sections/01-introduction}

\section{Related Work}
\label{sec:related_work}
\input{sections/02-related_work}

\section{Task Description}
\label{sec:task_description}
\input{sections/03-task_description}

\section{Method}
\label{sec:method}
\input{sections/04-method}

\section{Experiments}
\label{sec:experiments}
\input{sections/05-experiments}

\section{Conclusion}
\label{sec:conclusion}
\input{sections/06-conclusion}

\paragraph{Acknowledgements}
The authors would like to thank Steinar {\TH}orvaldsson for supporting the hardware demo and Vicente Ordonez for feedback on the manuscript.

%%%%%%%%% REFERENCES
{\small
\bibliographystyle{ieee_fullname}
\bibliography{main}
}

%\ifshowcomments  % hiding appendix when disabling comments
%    \clearpage
%    \section*{Appendix}
%    \label{sec:appendix}
%    \input{sections/appendix}
%\fi

\cleardoublepage
\twocolumn[%
{
\vskip .175in
\begin{center}
  {\Large \bf {\color{gray}(Supplementary Material)}\\ RREx-BoT: Remote Referring Expressions with a Bag of Tricks \par}
  \vspace*{24pt}
  \iftoggle{cvprfinal}{}{%
    \large
    \lineskip .5em
    \begin{tabular}[t]{c}
      Anonymous \confName~submission\\
      \vspace*{1pt}\\
      Paper ID \cvprPaperID
    \end{tabular}
    \par
    \vskip .5em
    \vspace*{12pt}
  }
\end{center}
}
]
\section{Appendix}
\label{sec:appendix}
\input{sections/appendix}

\end{document}

%% file: sections/01-introduction.tex
Household robots persist in the home over time, enabling them to build up knowledge of rooms, furniture, and objects.
For remote object grounding, an embodied agent is given a language description of a referent object, then conducts navigation and identification to localize that object.
We consider remote object grounding for agents that, like household robots, have familiarity with the environment.

\begin{figure}[ht]
    \centering
    \includegraphics[width=1.0\linewidth]{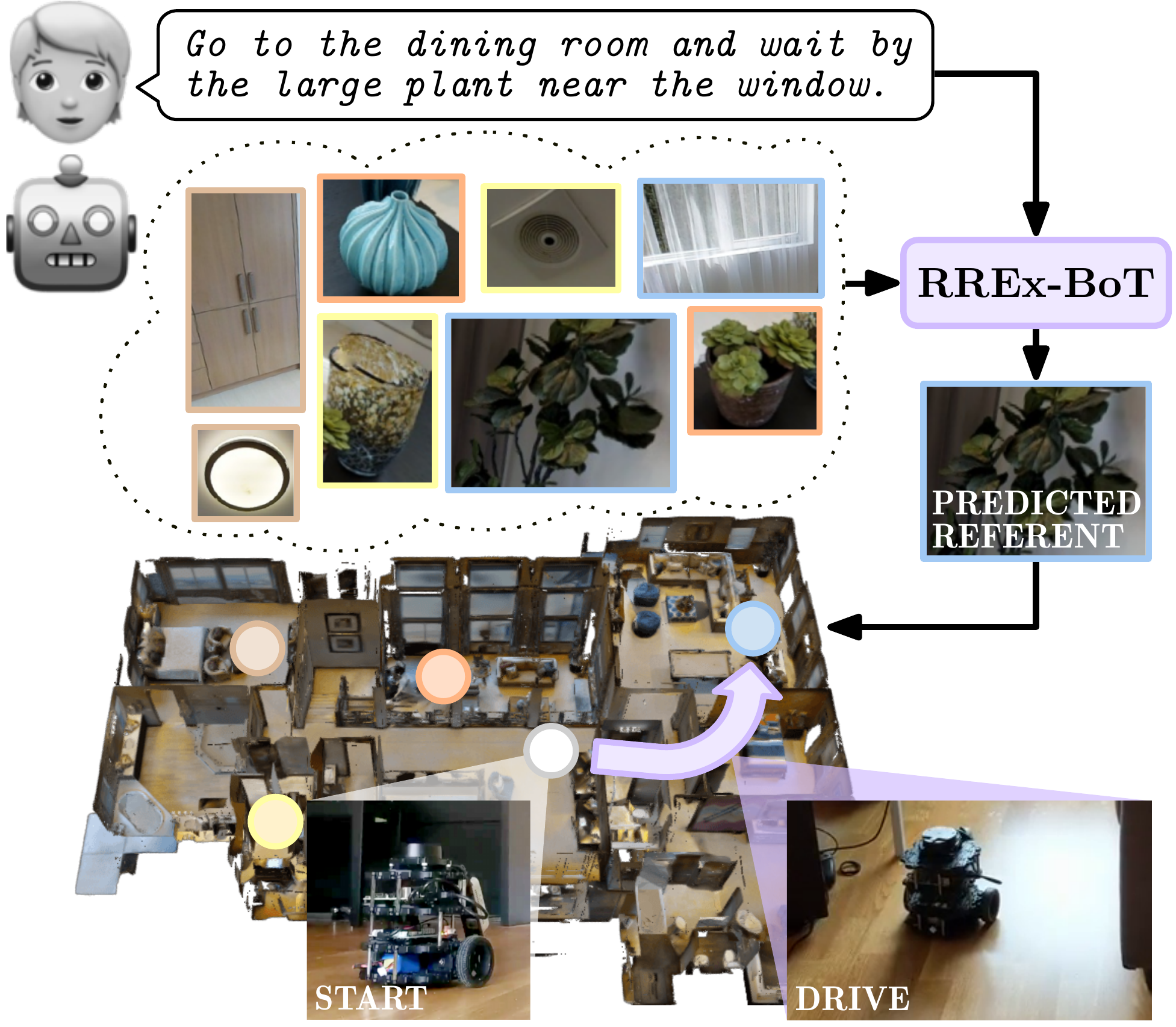}
    \caption{
    \ourmodel\ achieves state of the art performance on REVERIE and FAO by learning a ranking model on top of a generic vision-and-language backbone to score region proposals within a radius of an agent's starting position against a referring expression.
    The agent simply selects the highest scoring proposal as the referent and drives to it from the starting position.
    We deploy \ourmodel\ out-of-the-box to a physical robot platform for real world language-guided object navigation.
    }
    \label{fig:teaser}
\end{figure}

Remote object grounding approaches mostly assume an environment is \textit{totally unseen}.
However, methods for vision-and-language navigation frequently pre-explore environments~\cite{wang2020active}, or use beam search over trajectory rollouts at inference time~\cite{vlnbert,airbert}.
We bring exploration to the remote object grounding task.
Our problem formulation selects the best match for a referring expression from hundreds of thousands of objects in a environment (Figure~\ref{fig:teaser}).

With \ourmodelfull\ (\ourmodel), we set a new state of the art on remote object grounding without pre-exploration ($42.07\%$ versus the $32.23\%$ of AutoVLN~\cite{chen2022learning}) and with pre-exploration ($42.07\%$ versus the $34.69\%$ of OSMaN~\cite{cirik_thesis}) on the REVERIE~\cite{reverie} test set.
\ourmodel\ combines the strengths of two previous approaches.
\osman~\cite{cirik_thesis} provides an agent explicitly with a gold standard map and object region candidates.
The DUET model~\cite{duet}, by contrast, dynamically explores an unseen environment rather than taking advantage of a given map.
\ourmodel\ actively traverses the environment to learn a map, detecting object regions along the way, and identifies which detected object in the 3D environment most closely matches the provided language referring expression.

We formulate a ``bag of tricks'' that enable \ourmodel\ to \textit{efficiently} utilize a large-scale, pretrained language-and-vision alignment model as a backbone to score hundreds of thousands of objects against an input referring expression.
\begin{itemize}[noitemsep,topsep=0pt]
    \item We generalize from 2d to 3d positional embeddings for object proposals in an embodied environment.
    \item During training and inference we add context through additional automatically detected regions.
    \item During training, we augment training with viewpoints from which the referent region is not visible to regularize the model against false positives. 
    \item During inference, we explore viewpoints only up to a distance limit based on training data path lengths.
    \item During inference, we score region proposals in viewpoint groups with local neighborhood features to compare thousands of scores from hundreds of batches.
\end{itemize}
Figure~\ref{fig:main} outlines our training and inference procedure.

%% file: sections/02-related_work.tex
Remote object grounding is an instance of vision-and-language navigation (VLN), in which an embodied agent takes actions that move it through a simulated~\cite{macmahon:aaai06,chen:aaai11} or real environment~\cite{blukis2019learning} in response to a language instruction.
VLN tasks span from single agent navigation to two agent task completion~\cite{gu:acl22}.
Here, we overview dataset and modeling efforts in the subset of VLN regarding navigation towards \textit{object targets} based on language referring expressions.

\textbf{Benchmarks} for remote object grounding vary in the amount of visual information and language information given to an agent, as well as the nature of what is expected of the agent in response.
In 360-VQA, for example, an agent must answer a language question by examining a panoramic scene~\cite{360VQA}.
While the agent in 360-VQA does not need to take navigation actions, it must go beyond identifying a referent object and also answer a question about that object, for example \textquote{What color is the vase to the right of the pictures?}
By contrast, in Refer360$^{\circ}$ an agent adjusts its viewpoint, while still not performing navigation, in response to a \textit{series} of language instructions localizing an object before predicting a bounding box answer~\cite{refer360}.

Some works explore fidelity, continuous simulators~\cite{reve-ce,ivln}, utilize dialogue between two human players attempting to localize an agent in a scene based on object surroundings~\cite{way}, or navigate to a remote object known to one player but not another~\cite{cvdn}.
We develop our bag of tricks on a \textit{topological} simulator, where the agent only needs to reason over a discrete number of map locations, with single language instructions, rather than dialogue histories or interactions with oracles and simulated human partners.

Specifically, we focus on the Remote Embodied Visual Referring Expression in Real Indoor Environments (REVERIE)~\cite{reverie} and From Anywhere to Object (FAO) benchmarks~\cite{soon}.
We detail these in Section~\ref{sec:task_description}.

\textbf{Models} for remote object grounding include: iterative, sequence-to-sequence approaches; beam search methods to score potential trajectories; and those that maintain explicit environmental memory like maps.

Sequence-to-sequence approaches take in a language instruction to initialize a decoder that takes in image observations and emits actions~\cite{r2r}.
The History Aware Multimodal Transformer~\cite{chen2021hamt} is a successful sequence-to-sequence model which maintains a latent observations history used as additional conditioning information during action prediction.
A similar approach maintains only a single history state passed from one timestep to the next as a single token for an action-emitting transformer~\cite{cyclebert}.

Another class of models first gather rollouts from sequence-to-sequence models of \textit{possible} trajectories to follow from the start point conditioned on the language instruction.
Such beam search models are typically applied in vanilla VLN, not remote object grounding, due to VLN instructions being guiding, low-level path descriptions.
VLN-BERT is one such model, which rolls out possible paths based on a language-following model~\cite{fried2018speaker}, then feeds each candidate to a fine-tuned vision-action-language transformer model that scores how well the language instruction matches the observation sequence.
Airbert follows this same setup, but performs extensive domain-relevant fine tuning through data scraped from AirBnB~\cite{airbert}.

We first explore the environment to construct an explicit map memory with object region proposals at each location.
Such mapping-based approaches have been used for vision-and-language navigation via structured scene memory~\cite{Wang_2021_CVPR}, and for object localization in 3D environments via hierarchical mechanical search~\cite{kurenkov2020semantic}.
Our \ourmodel\ approach combines the strength of two existing mapping-based methods: DUET~\cite{duet} and \osman~\cite{cirik_thesis}.
In DUET, a topological map memory is built as the agent explores, and a combination of frontier exploration and instruction following is used to find the goal location.
In \osman, the agent is given the gold map and bounding box detections of all objects on initialization, breaking the assumptions of the benchmarks on which it is evaluated while enabling directly selecting a referent object from among all possibilities.
\ourmodel\ builds a map through exploration, as in DUET, but does so locally, recording object detections along the way to enable a final global scoring, as in \osman.

%% file: sections/03-task_description.tex
\begin{figure*}[ht]
    \centering
    \includegraphics[width=1.0\linewidth]{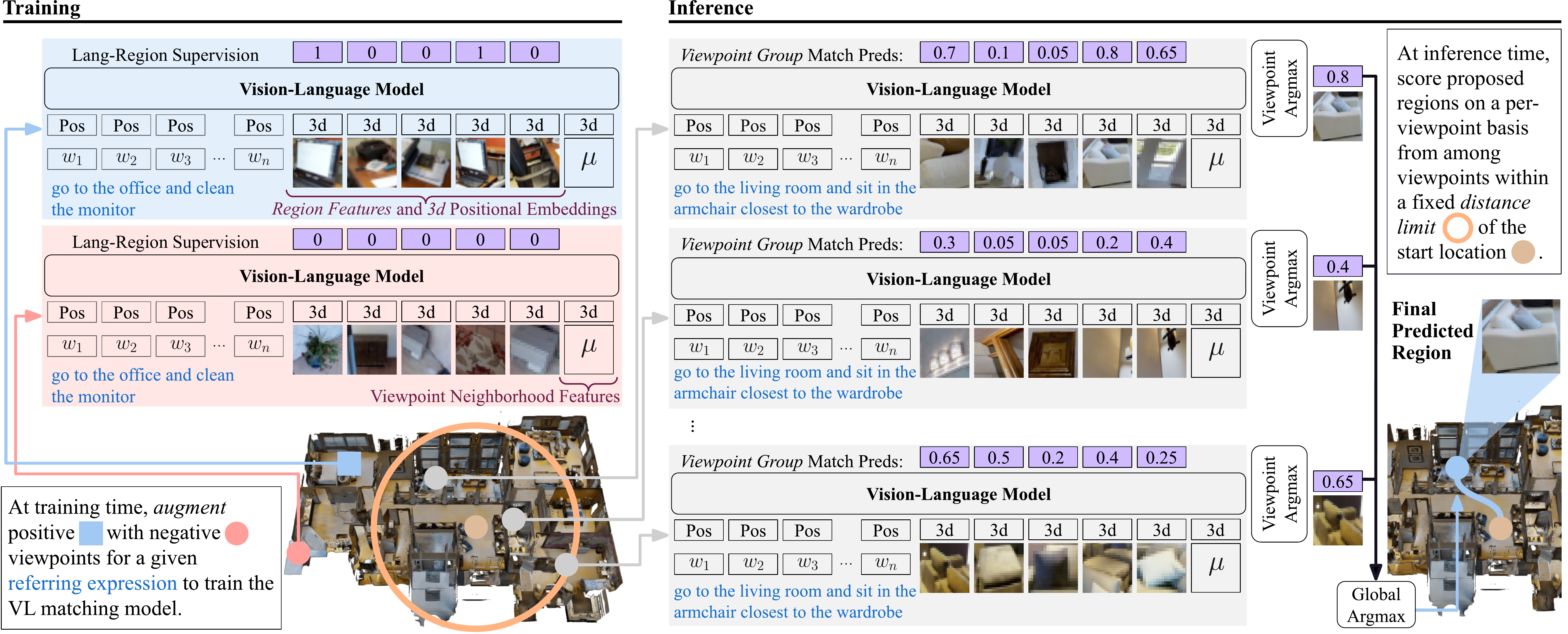}
    \caption{
    We achieve state of the art performance on remote object grounding after fully exploring an environment via a generic vision-language model with minor modifications and several changes to the training setup. 
    In particular, we start with a pretrained 2d vision-language backbone (ViLBERT~\cite{vilbert}) and fine-tune it for remote object grounding by adding 3d positional embeddings for region proposals.
    We train the backbone to score the matching between a referring expression and region proposals drawn from viewpoints in 3d space where the true target object can or cannot be seen.
    Then, at inference time, we explore up to a fixed number of steps away from a starting viewpoint, scoring regions on a per-viewpoint basis, then pool all best-scoring regions across viewpoints to find a global best within the exploration frontier, which we predict as the referent.
    }
    \label{fig:main}
\end{figure*}

Agents must navigate from a starting location and select the referent object based on region proposals at a goal location given a language referring expression.
Often, there are multiple goal positions from which the referent object is visible.
We consider the discrete Matterport3D~\cite{matterport} simulation environment used in the Remote Embodied Visual Referring Expression in Real Indoor Environments (REVERIE) task~\cite{reverie}.
Each navigable location is represented as a node in a navigation graph.
At each timestep, the agent receives a panoramic RGBD observation and a camera location and produces either a navigation action or an object targeting action to select the inferred referent object.
A panorama consists of images from any requested heading or elevation.

In the standard task formulation, the agent starts with no knowledge of the environment in each episode.
We consider an agent operating long-term in the same environment, a ``partially-observed'' version of this task, where an agent first explores the local environment up to a distance limit from the starting location and stores information before having to solve embodied referring expression tasks.

%% file: sections/04-method.tex
We face two major challenges when applying an off-the-shelf, generic vision-and-language alignment scoring model to the task of remote, embodied visual referring expression resolution.
First, such VL models are trained only on 2d images and associated language, but object referents in the current task are positioned in 3d space.
Second, these models output match scores between image proposals and target language in batches, where \textit{scores are relative within-batch} with respect to other batch regions given.
We cannot fit hundreds of thousands of object proposal regions in a single batch, making ranking language-image match scores between batches non-trivial.

% generic vision-language model
\subsection{Background: Vision-Language Modeling}
Our approach starts with a generic vision-language model formulation, where the visual input is represented with image region features. 
A region is a part of the image, such as a bounding box or object proposal, with corresponding feature and location information, typically obtained from an  object detector or an object proposal generator. 
We focus on the task where text input $\mathbf{w} {=} w_0, \dots, w_{T_w}$ is compared with a set of visual region features $\mathbf{v} {=} v_0, \dots, v_{T_v}$, where $T_w$ is the number of word tokens and $T_v$ image tokens. 
To encode token positional information, each token (text or visual) is paired with a positional encoding, capturing linear order for text or bounding box size and location for image regions.
This information is encoded in a latent feature and added to the corresponding token embedding.
Scoring networks are pre-trained on image and text description pairs to reconstruct masked out input tokens, learning a joint distribution between vision and language. 
They can then be fine-tuned on tasks such as choosing a corresponding visual region to given text input~\cite{Sharma2018,vilbert}.

\subsection{Bag-of-Tricks for Vision-Language Model in 3d}
Generic vision-and-language models are trained only on 2d image inputs.
There are two challenges for applying such models in large, 3d spaces: positional encodings of 3d space and an explosion in regions of interest.
Figure~\ref{fig:main} highlights our ``Bag of Tricks'' to overcome these challenges.

\paragraph{3d Positional Embeddings.}
To use the models in 3d, we replace the 2d bounding box coordinates $\left( x_0, y_0, x_1, y_1 \right)$ with 3d coordinates and size information $\left( x, y, z, r \right)$, where $r$ is the estimated radius of the object (half the diagonal of the bounding box in 3d coordinates). 
We normalize the $(x, y, z)$ location relative to the viewpoint $\pmb{p}$ from which the region was observed to calculate 3d positional encodings.

\paragraph{Region Scoring and Viewpoint Sampling.}
Pretrained models typically identify 10-100 regions of interest~\cite{vilbert} in a 2d image.
Thus, an embodied environment can have $10{,}000$-$100{,}000$ regions.\footnote{For example, a house with 250 viewpoints and 100 regions from 4 images in $360^\circ$, i.e. $250 \times 4 \times 100=100{,}000$}
Both training and inference, by contrast, proceed with small batches; in our case, 400 regions at a time can be processed.
Further, referring expressions often include information about the surrounding environment, and so region scores must be contextualized with local information.
Following an implementation detail in 2d vision-language models that add an average feature over all regions as a separate region~\cite{vilbert}, we add average features for each the 100 viewpoints closest to the viewpoint group (Figure~\ref{fig:main} \textit{Viewpoint Neighborhood Features}).

\ourmodel\ needs to be able to predict which regions are relevant per batch, including when \textit{none are relevant}.
We first sample regions from a single ground truth viewpoint---a location from which the target object is visible.
A sample can include zero or more ground truth regions, depending on how many of the bounding boxes cover the target object.
Since there is a large number of potential regions, and potential viewpoints, we found best to focus on regions from the ground truth viewpoint during training.
We use a sigmoid loss across these regions to train the model to select correct ground-truth regions while assigning low scores to distractor regions that are physically nearby:
\setlength{\abovedisplayskip}{3pt}
\setlength{\belowdisplayskip}{3pt}
\begin{align}
    \min_f \mathop{\mathbb{E}}_{\mathbf{w}, \mathbf{p}, \mathbf{y}} \left[ \sum_{v_i \in \mathbf{p}} \mathcal{L}(s_i, y_i) \right],
\end{align}
where $\mathbf{p} \subseteq \mathbf{v}$ is a viewpoint, i.e. a subset of visual regions, $s_i \in \mathbf{s}$ from $\mathbf{s} = f(\mathbf{p}, \mathbf{w})$ is the output sigmoid score for each region $v_i$ from the model $f$, and $y_i \in \mathbf{y}$ are ground truth labels for each region $v_i$ used with a binary cross-entropy loss $\mathcal{L}$.
To better calibrate the model to score all regions during inference we add augmentation on the viewpoint level (Figure~\ref{fig:main} \textit{Viewpoint Augmentation}). 
That is, during training we either select the ground truth viewpoint, or a random \textit{negative} viewpoint with a chance $R$.
Random viewpoints often contain no ground truth regions of the target object, and so training with these regularize the scores for incorrect viewpoints during inference.

\begin{figure}
    \centering
    \includegraphics[width=1.0\linewidth]{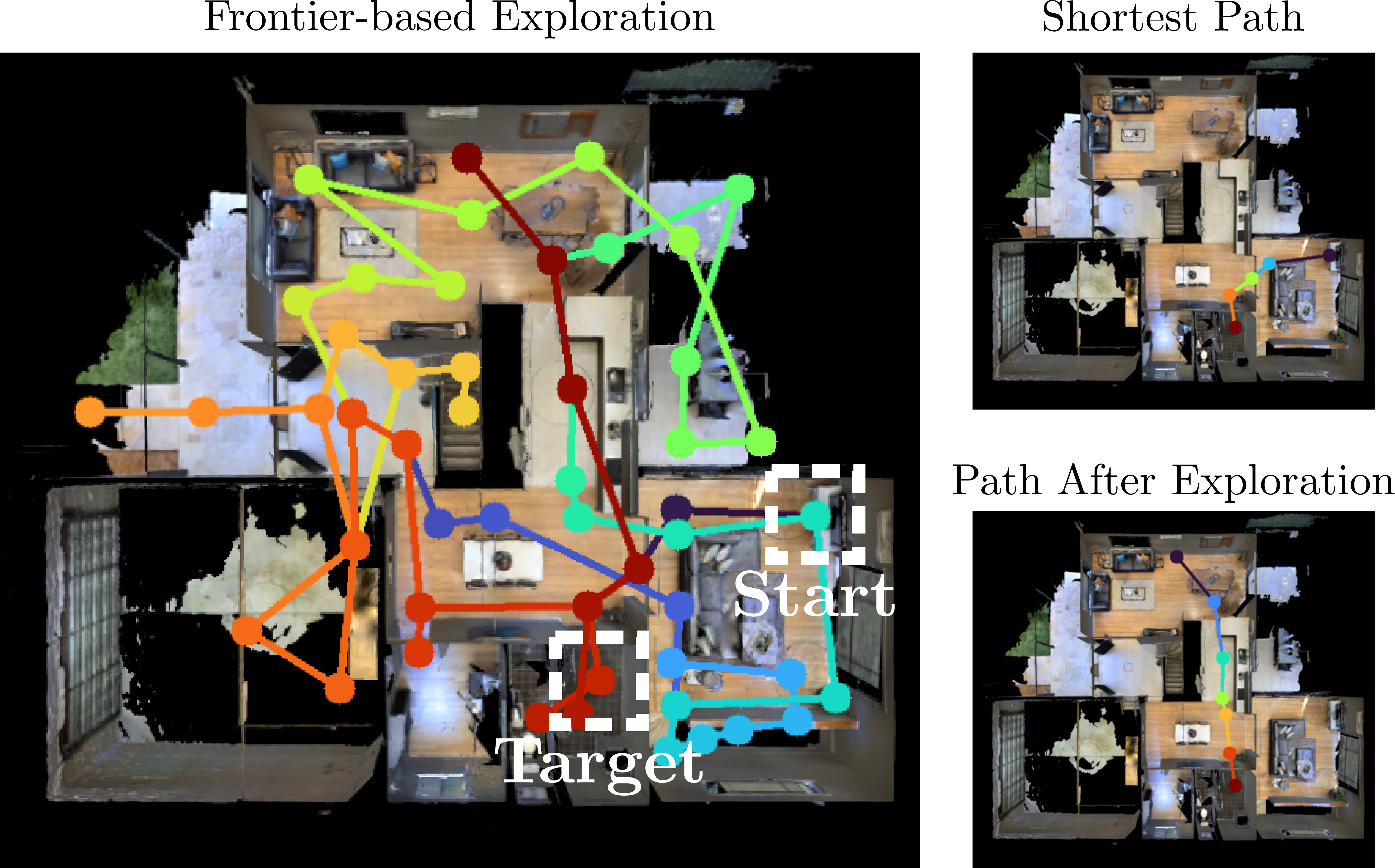}
    \caption{During inference, \ourmodel\ uses frontier-based exploration, or leverages a pre-explored map, then takes the shortest path to the target. 
    Here, we use blue to red colors for the start to end of a trajectory including exploration.
    }
    \label{fig:frontier}
\end{figure}

\paragraph{Non-Exhaustive Exploration}
We start each inference by performing frontier-based exploration~\cite{yamauchi1997frontier}.
However, to avoid scoring potential regions from all over the house, we cap exploration to up to $L$ steps from the starting point, with $L$ estimated based on the training set of each task as a reasonable upper bound on how far away target objects are from starting positions based on training data.
For a fair comparison with methods that do not pre-explore the house, the path the robot takes includes both the entire path taken by the frontier-based exploration, and then the estimated shortest path to the predicted destination. 
Figure~\ref{fig:frontier} illustrates the frontier-based exploration followed by the direct path to the goal, compared to the shortest path. 
For each panoramic viewpoint, we extract 400 region proposals, and use depth channel information to place each region proposal in 3d space.
The median depth value inside each region is used as the distance of the proposal from the camera origin. 

\input{tables/reverie_sota}

\paragraph{Inference-time Viewpoint Grouping}
To apply the trained model to $100{,}000$ regions at inference time, we batch regions based on the viewpoint from which they were extracted.
Thus, for a given inference text input ($w_0, \dots, w_{T_w}$), we process the the visual region features ($v_0, \dots, v_{T_v}$) from each viewpoint in separately (Figure~\ref{fig:main} \textit{Viewpoint Group Match Predictions}).
The scores are compared across viewpoints, and the highest scoring region is selected (Figure~\ref{fig:main} \textit{Viewpoint Argmax} and \textit{Global Argmax}).

%% file: tables/reverie_sota.tex
\begin{table*}[ht]
\footnotesize
\setlength{\aboverulesep}{0pt}
\setlength{\belowrulesep}{0pt}
\centering
\tabcolsep=0.07cm
\caption{Compared to state-of-the-art methods on the REVERIE dataset, \ourmodel\ outperforms all in Remote Grounding Success (RGS) and navigation success rate (SR).
With pre-exploration, \ourmodel\ also outperforms alternative pre-exploration methods on these metrics as well as their path-weighted variants (RGSPL, SPL).
Test Unseen results are from the public leaderboard~\cite{reverie_leaderboard}. 
Table adapted from~\cite{duet}.
}
\label{tab:reverie}
\begin{tabular}{@{}l@{}rrrroorrrroorrrroo} 
& \multicolumn{6}{c}{\it Val Seen} & \multicolumn{6}{c}{\it Val Unseen} & \multicolumn{6}{c}{\it Test Unseen} \\
& \multicolumn{4}{c}{Navigation} & \multicolumn{2}{k}{Grounding} & \multicolumn{4}{c}{Navigation} & \multicolumn{2}{k}{Grounding} & \multicolumn{4}{c}{Navigation} & \multicolumn{2}{k}{Grounding}  \\ 
\bf Method & TL$\downarrow$ & OSR$\uparrow$ & SR$\uparrow$ & SPL$\uparrow$ & RGS$\uparrow$ & {\scriptsize RGSPL}$\uparrow$ & TL$\downarrow$ & OSR$\uparrow$ & SR$\uparrow$ & SPL$\uparrow$ & RGS$\uparrow$ & {\scriptsize RGSPL}$\uparrow$ & TL$\downarrow$ & OSR$\uparrow$ & SR$\uparrow$ & SPL$\uparrow$ & RGS$\uparrow$ & {\scriptsize RGSPL}$\uparrow$ \\ 
\toprule
Human & - & - & - & - & - & - & - & - & - & - & - & - & $21.18$ & $86.83$ & $81.51$ & $53.66$ & $77.84$ & $51.44$ \\ \midrule
\scriptsize Seq2Seq\cite{r2r} & $12.88$ & $35.70$ & $29.59$ & $24.01$ & $18.97$ & $14.96$ & $11.07$ & $8.07$ & $4.20$ & $2.84$ & $2.16$ & $1.63$ & $10.89$ & $6.88$ & $3.99$ & $3.09$ & $2.00$ & $1.58$ \\
\scriptsize RCM\cite{wang2019reinforced} & $10.70$ & $29.44$ & $23.33$ & $21.82$ & $16.23$ & $15.36$ & $11.98$ & $14.23$ & $9.29$ & $6.97$ & $4.89$ & $3.89$ & $10.60$ & $11.68$ & $7.84$ & $6.67$ & $3.67$ & $3.14$ \\
\scriptsize SMNA\cite{ma2019self} & $7.54$ & $43.29$ & $41.25$ & $39.61$ & $30.07$ & $28.98$ & $9.07$ & $11.28$ & $8.15$ & $6.44$ & $4.54$ & $3.61$ & $9.23$ & $8.39$ & $5.80$ & $4.53$ & $3.10$ & $2.39$ \\
\scriptsize FAST-MATTN\cite{Ke2019TacticalRS} & $16.35$ & $55.17$ & $50.53$ & $45.50$ & $31.97$ & $29.66$ & $45.28$ & $28.20$ & $14.40$ & $7.19$ & $7.84$ & $4.67$ & $39.05$ & $30.63$ & $19.88$ & $11.61$ & $11.28$ & $6.08$ \\
\scriptsize SCoA\cite{zhu2021self} & - & - & - & - & - & - & - & $29.29$ & $16.94$ & $8.20$ & - & - & - & - & - & - & - & - \\
\scriptsize CKR\cite{Gao2021RoomandObjectAK} & $12.16$ & $61.91$ & $57.27$ & $53.57$ & $39.07$ & - & $26.26$ & $31.44$ & $19.14$ & $11.84$ & $11.45$ & - & $22.46$ & $30.40$ & $22.00$ & $14.25$ & $11.60$ & - \\
\scriptsize SIA\cite{lin2021scene} & $13.61$ & $65.85$ & $61.91$ & $57.08$ & $45.96$ & $42.65$ & $41.53$ & $44.67$ & $31.53$ & $16.28$ & $22.41$ & $11.56$ & $48.61$ & $44.56$ & $30.80$ & $14.85$ & $19.02$ & $9.20$ \\
\scriptsize ORIST\cite{orist} & $10.73$ & $49.12$ & $45.19$ & $42.21$ & $29.98$ & $27.77$ & $10.90$ & $25.02$ & $16.84$ & $15.14$ & $8.52$ & $7.58$ & $11.38$ & $29.20$ & $22.19$ & $18.97$ & $10.68$ & $9.28$ \\
\scriptsize ProbES\cite{liang2022visual} & $13.59$ & $48.49$ & $46.52$ & $42.44$ & $33.66$ & $30.86$ & $18.00$ & $33.23$ & $27.63$ & $22.75$ & $16.84$ & $13.94$ & $17.43$ & $28.23$ & $24.97$ & $20.12$ & $15.11$ & $12.32$ \\
\scriptsize RecBERT\cite{cyclebert} & $13.44$ & $53.90$ & $51.79$ & $47.96$ & $38.23$ & $35.61$ & $16.78$ & $35.02$ & $30.67$ & $24.90$ & $18.77$ & $15.27$ & $15.86$ & $32.91$ & $29.61$ & $23.99$ & $16.50$ & $13.51$ \\
\scriptsize Airbert\cite{airbert} & $15.16$ & $48.98$ & $47.01$ & $42.34$ & $32.75$ & $30.01$ & $18.71$ & $34.51$ & $27.89$ & $21.88$ & $18.23$ & $14.18$ & $17.91$ & $34.20$ & $30.28$ & $23.61$  & $16.83$ & $13.28$ \\
\scriptsize HAMT\cite{chen2021hamt} & $12.79$ & $47.65$ & $43.29$ & $40.19$ & $27.20$ & $25.18$ & $14.08$ & $36.84$ & $32.95$ & $30.20$ & $18.92$ & $17.28$ & $13.62$ & $33.41$ & $30.40$ & $26.67$ & $14.88$ & $13.08$ \\
\scriptsize HOP\cite{Qiao2022HOP} & $13.80$ & $54.88$ & $53.76$ & $47.19$ & $38.65$ & $33.85$ & $16.46$ & $36.24$ & $31.78$ & $26.11$ & $18.85$ & $15.73$ & $16.38$ & $33.06$ & $30.17$ & $24.34$ & $17.69$ & $14.34$ \\
\scriptsize TD-STP\cite{zhao2022target} & - & - & - & - & - & - & - & $39.48$ & $34.88$ & $27.32$ & $21.16$ & $16.56$ & - & $40.26$ & $35.89$ & $27.51$ & $19.88$ & $15.40$ \\
\scriptsize RecBERT(O)\cite{ZHAN202368} & $14.40$ & $67.81$ & $65.50$ & $60.12$ & $48.28$ & $43.95$ & $22.20$ & $40.24$ & $32.58$ & $25.90$ & $22.01$ & $17.52$ & $22.91$ & $44.44$ & $36.13$ & $26.95$ & $22.08$ & $16.75$ \\
\scriptsize DUET\cite{duet} & $13.86$ & {$73.86$} & {$71.75$} & {$\pmb{63.94}$} & {$57.41$} & {$\pmb{51.14}$} & $22.11$ & {$51.07$} & {$46.98$} & {$33.73$} & {$32.15$} & {$23.03$} & $21.30$ & {$56.91$} & {$52.51$} & {$36.06$} & {$31.88$} & {$22.06$} \\
\scriptsize AutoVLN\cite{chen2022learning} & - & - & - & - & - & - & - & $62.14$ & $55.89$ & $\pmb{40.85}$ & $36.58$ & $\pmb{26.76}$ & - & $62.30$ & $55.17$ & $\pmb{38.88}$ & $32.23$ & $\pmb{22.68}$ \\
\ourmodel & $212.6$ & $100.0$ & $\pmb{77.37}$ & $4.69$ & $\pmb{62.97}$ & $3.71$ & $185.6$ & $100.0$ & $\pmb{61.06}$ & $3.53$ & $\pmb{43.74}$ & $2.47$ & $177.2$ & $100.0$ & $\pmb{65.18}$ & $4.34$ & $\pmb{42.07}$ & $2.78$ \\
\midrule
\multicolumn{10}{l}{\it ---Methods Utilizing Pre-Exploration---} \\
\scriptsize OSMaN\cite{cirik_thesis} & - & - & $60.9$ & $58.9$ & $43.5$ & $41.8$ & - & - & $53.40$ & $51.50$ & $36.00$ & $34.60$ & $8.67$ & $57.12$ & $50.16$ & $47.78$ & $34.69$ & $32.99$ \\
\ourmodel(PE) & $10.65$ & $79.83$ & $\pmb{77.30}$ & $\pmb{75.95}$ & $\pmb{62.97}$ & $\pmb{61.81}$ & $9.54$ & $68.56$ & $\pmb{61.01}$ & $\pmb{58.83}$ & $\pmb{43.68}$ & $\pmb{42.24}$ & $9.57$ & $73.74$ & $\pmb{65.19}$ & $\pmb{62.04}$ & $\pmb{42.05}$ & $\pmb{40.12}$ \\
\bottomrule
\end{tabular}
\end{table*}

%% file: sections/05-experiments.tex
We demonstrate how a simple vision-language model can achieve state of the art results on the VLN benchmarks REVERIE~\cite{reverie} and FAO~\cite{soon}. 
Highlighting the simplicity and usefulness, the supplementary material includes a video demonstration of this system running on a real-world TurtleBot3 platform by first pre-exploring an environment and then navigating to objects from a textual description.

\subsection{Benchmarks and Metrics}
\label{sec:experiments:datasets}
We conduct experiments by using simulator-based datasets REVERIE~\cite{reverie} and FAO~\cite{soon}.
REVERIE contains 10{,}466 training instructions that contain 21 words on average, and the ground truth trajectories are 4-7 steps. 
Agent observations are 640x480 pixels with $60^\circ$ vertical field-of-view ($80^\circ$ horizontal field-of-view).
FAO\footnote{\url{https://scenario-oriented-object-navigation.github.io/} also known as the SOON dataset.} 
contains 26{,}790 training instructions that contain 47 words on average, and the ground truth trajectories are 2-21 steps (9.5 on average). We use the proposal bounding boxes provided in DUET~\cite{duet} to improve the training data. 
As in DUET, we use observations of 600x600 pixels with $80^\circ$ vertical and horizontal field-of-view, making our models directly comparable.

\paragraph{Evaluation Metrics}
For navigation, we consider average path length in meters (\emph{Trajectory Length, TL}), fraction of paths ending where the target is visible (REVERIE) or within 3 meters of it (FAO) (\emph{Success Rate, SR}), best SR achieved along the path (\emph{Oracle Success Rate, OSR}), and \emph{SR} penalized by path length (\emph{SPL}). 
For object grounding, we consider the episodes choosing the correct object (\emph{Remote Grounding Succcess, RGS}), and \emph{RGS} penalized by path length (\emph{RGSPL}).\footnote{In FAO~\cite{soon}, RGSPL is referred to as SFPL; we follow DUET~\cite{duet} and use the more common RGSPL name.}
Grounding is successful if the chosen region bounding box has over 50\% IoU with the ground truth object and is selected from a valid goal viewpoint.

\input{tables/fao_sota.tex}

\subsection{Implementation Details}
At a high level, \ourmodel\ first visits all locations in a given scene, builds a list of proposed or observed regions of interest, and then uses that list to select the best referent region for a given referring expression.
We fine-tune the ViLBERT vision-language model~\cite{vilbert}, which is pre-trained on the Contextual Captions~\cite{Sharma2018} dataset, to score how well referring expressions match proposed regions.
We choose ViLBERT because it is a generic VL backbone model with no task-specific architecture or objectives tuned for sequential decision making or the VLN setting. 

We exclude all viewpoints that are more than $L$ steps away from the starting viewpoint location of the agent in each episode. 
We set $L{=}8$ for REVERIE, the maximum length trajectory in training, and $L{=}13$ for FAO, based on a validation performance.

We use the region proposals provided by REVERIE itself for that setting, and by DUET~\cite{duet} for FAO.
For training, we augment these regions with new proposals predicted by Mask-RCNN~\cite{maskrcnn} collected at headings $0^\circ$, $90^\circ$, $180^\circ$, $270^\circ$ at $0^\circ$ elevation to add additional contextual information to reach $T_v{=}400$ total.
At inference we choose among only the provided proposals to select the final grounding.

We train for 160 epochs on REVERIE and 30 epochs on FAO, using a batch size of 10. We use a learning rate of 1e-5, weight decay of 0.01, and a 10\%/90\% linear warmup/decay schedule. We set the viewpoint augmentation chance to $R{=}80\%$, meaning 80\% of training viewpoints could contain no correct referent regions (Figure~\ref{fig:main} \textit{Training}). 
Each model was trained on a single NVIDIA T4 GPU, and training took 4 days.
Source code for replicating the results in this paper will be made available. 

\subsection{Benchmark Results}
\ourmodel\ outperforms state of the art methods on REVERIE and FAO, with and without pre-exploration, and is easily deployed to a real robot system.

\paragraph{REVERIE Results}
Table~\ref{tab:reverie} compares \ourmodel\ to many others. 
At inference time, \ourmodel\ first explores all viewpoints within $L$ steps before scoring regions and navigating to the predicted referent.
We also include a \ourmodel\ variant that pre-explores (PE) the environments and so can navigate directly to the selected referent, similar to the OSMaN~\cite{cirik_thesis} model.
The \textit{Test Unseen} results are obtained from the public leaderboard, where our \ourmodel\ is, at the time of writing, state-of-the-art in terms of \emph{SR} and \emph{RGS}, and \emph{\ourmodel\ Pre-Explore} is state-of-the-art in terms of \emph{SPL} and \emph{RGSPL} among pre-exploration models~\cite{reverie_leaderboard}.

Base \ourmodel\ achieves low \emph{SPL} and \emph{RGSPL} metrics that consider path length due to frontier exploration.
The oracle success rate \emph{OSR} is $100.0\%$ because all locations where the referent \textit{can be} are a part of the exploration path with $L{=}8$.
\emph{\ourmodel\ (PE)} evaluations when exploration is not included as part of the path, similar to OSMaN~\cite{cirik_thesis}.
We exceed OSMaN on path weighted \emph{SPL} and \emph{RGSPL} metrics, demonstrating that in cases where the agent has been operating in the same environment continuously, a simple model can significantly outperform specialized models.

\paragraph{FAO Results}
Table~\ref{tab:fao} shows results on the unseen validation set and the held-out challenge test set of the FAO dataset~\cite{soon}.\footnote{The regular remote grounding success (RGS) metric, without path penalty, was not presented in the original FAO paper~\cite{soon}, or on the leaderboard. Running the model evaluation from the DUET~\cite{duet} open-source code included that number however, and that is included here.}
As with REVERIE, \ourmodel\ achieves state of the art results. 
The FAO dataset includes longer paths than REVERIE and noisier, more challenging training data, but \ourmodel\ adapts to this setting and reach a convincing performance.
The oracle success rate \emph{OSR} is lower than $100\%$ because our exploration budget of $L{=}13$ is lower than that needed to reach all targets.
Setting $L{=}21$ would reach $100\%$ OSR, but $L{=}13$ yields better performance because of a more tractable search space over regions.

\paragraph{Home Robot Results}
We evaluate \ourmodel\ on a TurtleBot3 Burger with an attached RGBD camera running ROS nodes for coordinate frames, navigation, and camera updates.
The robot first drives around the environment and running Mask-RCNN on RGBD images to create regions (which have 3d coordinates from the depth).
Then, we use \ourmodel\ to select a region proposal from among those stored given a text query.
The selected regions' 3d coordinates are passed to the ROS navigation system. 
The robot was able to build a map of the environment, use \ourmodel\ to identify the correct object based on a query, and use the navigation system to navigate to that target.
In the supplementary material we include a video demonstration of the robot performing mapping, remote grounding, and successful navigation to a referent target given speech input.

\begin{figure*}[ht]
    \centering
    \includegraphics[width=1.0\linewidth]{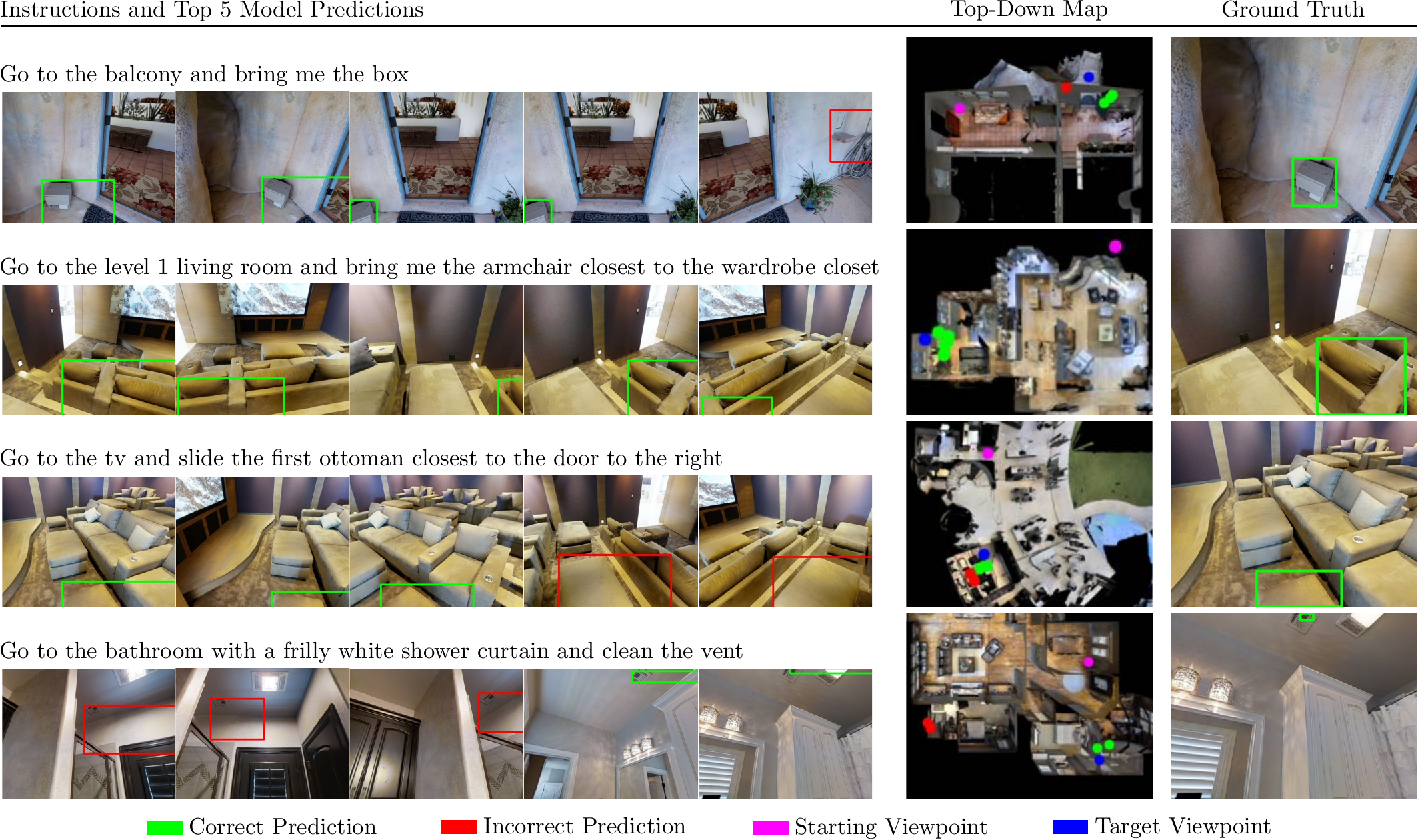}
    \caption{Qualitative results from our model. The visualizations show a view of the ground truth target object given the text instructions, along with the top 5 predictions from the model from left to right. We also show a top-down map of the floor containing the target object with the starting location (pink), goal location (blue), correct predictions (green), incorrect predictions (red). 
    The top 3 rows are successful remote grounding (top 1 predicted object is correct) whereas the last row is unsuccessful (top 3 predictions incorrect).}
    \label{fig:qualitative}
\end{figure*}

\begin{figure}[ht]
    \centering
    \includegraphics[width=1.0\linewidth]{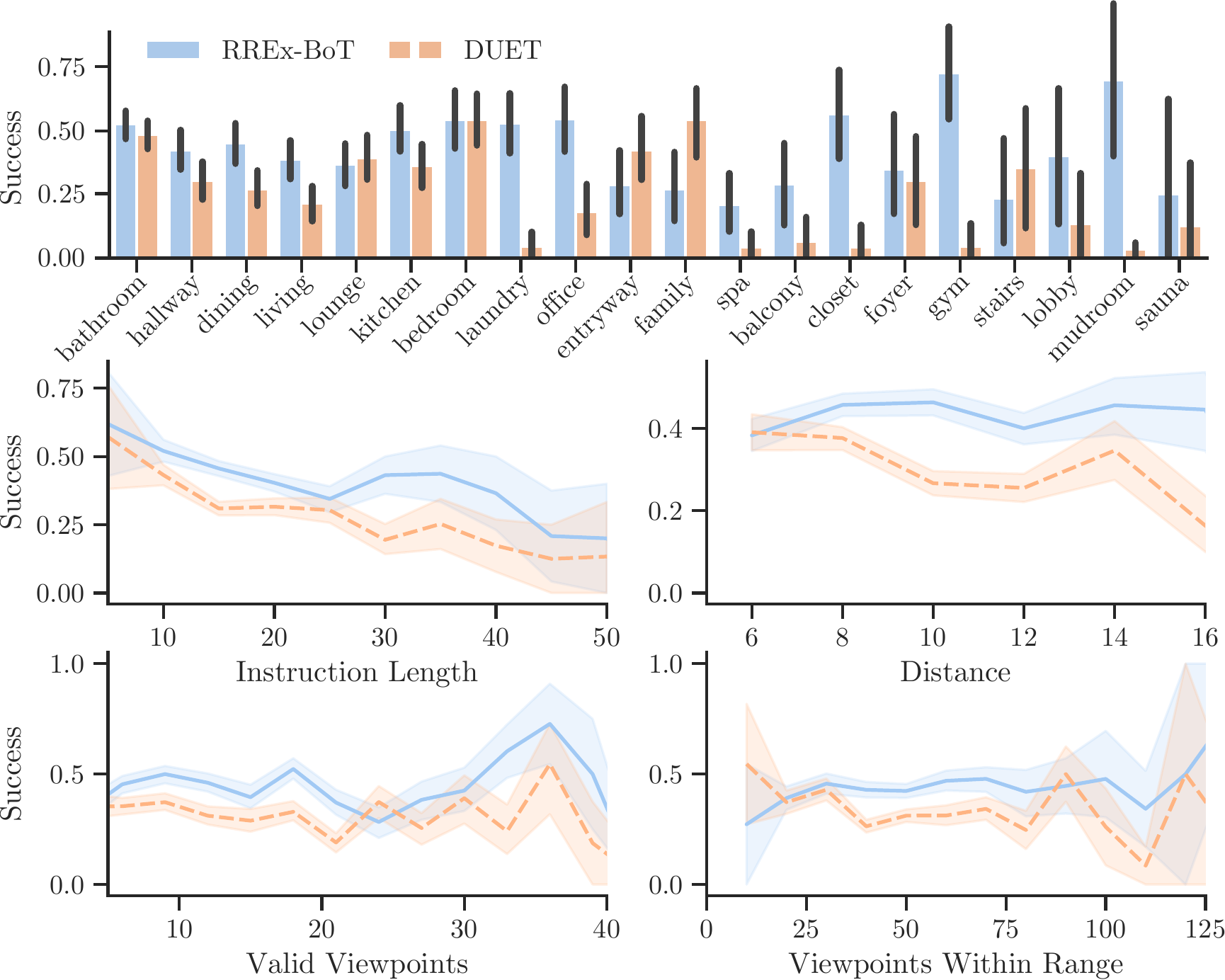}
    \caption{Analysis of RGS (\emph{Success}) of \ourmodel \ on REVERIE unseen validation. Success is plotted as a function of various parameters. The shaded region is the 95\% bootstrapping confidence interval of the estimate, where small sample sizes result in larger error bounds. DUET~\cite{duet} is included as a baseline.}
    \label{fig:analysis}
\end{figure}

\subsection{Performance Analysis and Ablations}
Figure~\ref{fig:qualitative} shows correct and incorrect \ourmodel\ predictions in REVERIE.
To better understand where \ourmodel\ does and does not work well, we analyze its performance on REVERIE in detail, including ablations of various aspects of each entry in our ``bag of tricks.''

\paragraph{Analysis on REVERIE} Highlighting the challenge for generalizing an image vision-language model to 100{,}000s of regions, we found that if we limit evaluation to only regions from a viewpoint where the goal is visible,\footnote{Corresponding to the training scenario without augmentation.} \ourmodel\ can reach a $98.5\%$ RGS on the training set. 
In contrast, this performance is $78.7\%$ on the validation seen, and $62.9\%$ on the validation unseen set.

We found that $30.8\%$ of the errors on the unseen validation were from a valid viewpoint (versus $20.2\%$ in DUET), and if a trajectory ended in a valid viewpoint there was a $71.6\%$ chance it would have remote grounding success (versus $68.7\%$ in DUET).
This difference demonstrates the advantage of the global search after pre-exploration, since there are fewer failures from invalid viewpoints. 

Figure~\ref{fig:analysis} compares \ourmodel\ with DUET~\cite{duet} on the \textit{Validation Unseen} split.
We look at success as a function of: the room type of the target object; number of words in the referring expression (\emph{Instruction Length}); shortest distance to target in meters (\emph{Distance}); number of viewpoints where the ground truth object can be found (\emph{Valid Viewpoints}); number of viewpoints within $L{=}8$ distance from the starting viewpoint (\emph{Viewpoints Within Range}).
We find that \ourmodel\ is relatively robust to the distance and size of the search space, but could be improved with more robust language understanding for longer language inputs.

\input{tables/ablations}

\paragraph{Ablation Studies}
\label{sec:ablations}
We consider implementation choices for each ``trick'' used by \ourmodel.
We examine the pre-exploration setting to study upper bound of \ourmodel\ with ablated components.
Table~\ref{tab:ablations:overview} evaluates \ourmodel (PE): with no 3d information (\emph{-- 3D Positional Encoding}), with no additional region proposals for additional context (\emph{-- Context Proposals}), with all regions considered (\emph{-- Distance Limit}), without any viewpoints that do not contain the target during training (\emph{-- Augmentation}), with regions processed in random batches instead of by viewpoint (\emph{-- Viewpoint Grouping}), and using pre-trained ViLBERT as-is without any fine-tuning on the REVERIE data (\emph{-- Fine-Tuning}).

We find that grouping by viewpoint is critical to simplify the search problem, and augmentation with viewpoints that may not contain referent regions is important to suppress false positives when scoring 100{,}000s of proposal regions. 
Since the evaluation setup for REVERIE and FAO includes a start position, and the trajectories do not typically cover the entire house, it is also important to include a budget on the search space. 
There are noticeable, if less dramatic, improvements when using 3d positional encoding and when adding additional in-context proposals from an object detector.
We investigate each of these findings in more detail.

\textit{\textbf{How to find needles in haystacks?}}
Bridging the gap between the training setup (400 regions) and the final inference setup ($100{,}000$ regions) requires carefully managing overfitting of the model. 
In Table~\ref{tab:ablations:specific} (\textit{Aug}) we first vary the viewpoint augmentation and see that decreasing to 50\% (from \ourmodel's 80\%) loses performance, while increasing to 90\% has a minor negative effect.
In Table~\ref{tab:ablations:specific} (\textit{Regions}), we consider sampling with replacement the regions in each viewpoint during training (\emph{+ Bootstrapping}), and environment dropout~\cite{tan-etal-2019-learning} where we randomly replace regions during training with other regions from the training set (\emph{+ Env. Dropout 50/75\%}), which seem to not help in this case.

\textit{\textbf{How to encode 3d positions?}}
We explore different ways of encoding the 3d information. 
In Table~\ref{tab:ablations:specific} (\textit{3d Enc}) we show result for using the bounding box coordinates in the panorama space (\emph{BBox Coord.}), absolute 3d coordinates (\emph{Absolute Coord.}), 3d coordinates relative to the start of the trajectory (\emph{Start Relative Coord.}), and constant 3d coordinates unrelated to the region (\emph{No Region Positional Enc}). 
The different types of 3d information perform similarly, with \emph{Viewpoint Rel. Coord.}, used by \ourmodel, performing slightly better.

\textit{\textbf{How to add 3d context?}}
Investigating how to incorporate 3d context (surrounding information) when making a prediction we look at the following settings in Table~\ref{tab:ablations:specific} (\textit{3dC}): no additional object proposals (\emph{-- Context Proposals}), and no average features from neighboring viewpoints (\emph{-- Viewpoint Nbhd. Feat.}). 
While having additional object proposals is important, we found additional context less useful.

\textit{\textbf{How to reduce the 3d search space?}}
Finally, we consider different ways of reducing the search space during inference, since $100{,}000$ regions can create many false positives.
In Table~\ref{tab:ablations:specific} (\textit{Infer}) we note that removing the \emph{Viewpoint Grouping} and \emph{Distance Limit} has a drastic effect on the final model. 
However, adding a \emph{Two-Step Inference} approach, were we first classify among the different viewpoints (average feature for each) and then between the regions in best viewpoint, reduces the search space too much, and causes a slight drop in performance.
Interestingly, we found that including 3d positional encoding was not enough for the network to learn the ``distance limit'' and ignore everything beyond a certain threshold. 
This result is likely because a constraint of $L$ steps in the simulator does not translate well into Euclidean distance in 3d space.

%% file: tables/fao_sota.tex
\begin{table*}[ht]
\setlength{\aboverulesep}{0pt}
\setlength{\belowrulesep}{0pt}
\tabcolsep=0.07cm
\centering
\small
\caption{Compared to state of the art on the FAO dataset~\cite{soon}, \ourmodel\ outperforms all in grounding (RGS) and navigation (SR) metrics.
With pre-exploration (PE), the path weighted RGSPL and SPL similarly improve, but there are no other PE model results against which to compare for FAO. Test Unseen results are from the public leaderboard~\cite{soon_leaderboard}. 
}
\label{tab:fao}
\begin{tabular}{@{}l@{}rrrroorrrroo}
& \multicolumn{6}{c}{\it Val Unseen} & \multicolumn{6}{c}{\it Test Unseen} \\
& \multicolumn{4}{c}{Navigation} & \multicolumn{2}{k}{Grounding} & \multicolumn{4}{c}{Navigation} & \multicolumn{2}{k}{Grounding} \\ 
\bf Method & TL$\downarrow$ & OSR$\uparrow$ & SR$\uparrow$ & SPL$\uparrow$ & RGS$\uparrow$ & RGSPL$\uparrow$ & TL$\downarrow$ & OSR$\uparrow$ & SR$\uparrow$ & SPL$\uparrow$ & RGS$\uparrow$ & RGSPL$\uparrow$ \\ 
\toprule
GBE\cite{soon} & $28.96$ & $28.54$ & $19.52$ & $13.34$ & - & $1.16$ & $27.88$ & $21.45$ & $12.90$ & $9.23$ & - & $0.45$ \\
DUET\cite{duet} & $36.20$ & $50.91$ & $36.28$ & $22.58$ & $6.02$ & $3.75$ & $41.83$ & $43.00$ & $33.44$ & $21.42$ & - & $4.17$ \\
AutoVLN\cite{chen2022learning} & - & $53.19$ & $41.00$ & $\pmb{30.69}$ & - & $\pmb{4.06}$ & - & $48.74$ & $40.36$ & $\pmb{27.83}$ & - & \pmb{$5.11$} \\
\ourmodel & $250.4$ & $97.37$ & $\pmb{49.17}$ & $3.65$ & $\pmb{11.06}$ & $0.83$ & $339.3$ & $98.16$ & $\pmb{47.53}$ & $3.34$ & - & $0.64$ \\
\midrule
\ourmodel(PE) & $15.60$ & $56.40$ & $\pmb{49.17}$ & $\pmb{48.63}$ & $\pmb{11.06}$ & $\pmb{11.06}$ & $15.38$ & $52.07$ & $\pmb{47.48}$ & $\pmb{47.13}$ & - & $\pmb{9.89}$ \\
\bottomrule
\end{tabular}
\end{table*}

%% file: tables/ablations.tex
\begin{table}[ht]
\setlength{\aboverulesep}{0pt}
\setlength{\belowrulesep}{0pt}
\tabcolsep=0.07cm
\centering
\small
\caption{Ablations of \ourmodel\ ``tricks'' on REVERIE.}
\label{tab:ablations:overview}
\begin{tabular}{lroro} 
    & \multicolumn{2}{c}{\it Val Seen} & \multicolumn{2}{c}{\it Val Unseen} \\
    \bf Method & SR$\uparrow$ & RGS$\uparrow$ & SR$\uparrow$ & RGS$\uparrow$ \\ 
    \toprule
    \ourmodel & $\pmb{77.37}$ & $\pmb{62.97}$ & $\pmb{61.06}$ & $\pmb{43.75}$ \\
    \midrule
    -- Region Positional Enc. & $75.90$ & $61.07$ & $58.90$ & $40.50$ \\
    -- Context Proposals & $74.63$ & $59.94$ & $52.49$ & $35.64$ \\
    -- Distance Limit & $72.17$ & $59.24$ & $45.38$ & $31.50$ \\
    -- Augmentation & $30.15$ & $27.62$ & $16.39$ & $14.14$ \\
    -- Viewpoint Grouping & $33.59$ & $13.77$ & $23.69$ & $10.17$ \\
    -- Fine-Tuning & $10.89$ & $\phantom{0}2.60$ & $12.67$ & $\phantom{0}4.23$ \\
    \bottomrule
\end{tabular}
\end{table}

\begin{table}[ht]
\setlength{\aboverulesep}{0pt}
\setlength{\belowrulesep}{0pt}
\tabcolsep=0.07cm
\centering
\small
\caption{\ourmodel\ performance on REVERIE unseen validation set as we adjust hyperparameters and ablate components within each ``trick.''
We examine the negative viewpoint augmentation rate (Aug; 80\% used by selected model), region sampling strategies (Regions), replacing our Viewpoint Relative Coordinates with different kinds of 3d positional encodings for regions (3d Enc), removing 3d context information input to ViLBERT (3dC), and changing how we group regions and limit viewpoints considered during inference (Infer).
}
\label{tab:ablations:specific}
\begin{tabular}{llro} 
    & \bf Method Ablation & SR$\uparrow$ & RGS$\uparrow$ \\ 
    \toprule
        & \ourmodel & $\pmb{61.06}$ & $\pmb{43.75}$ \\
    \midrule
        \multirow{2}{*}{\rotatebox[origin=c]{90}{Aug}}
        & 50\% Negative Viewpoints & $57.71$ & $39.48$ \\
        & 90\% Negative Viewpoints & $61.06$ & $42.66$ \\
    \midrule
        \multirow{3}{*}{\rotatebox[origin=c]{90}{Regions}}
        & + Bootstrapping & $60.72$ & $41.18$ \\
        & + Env. Dropout 50\% & $56.49$ & $39.05$ \\
        & + Env. Dropout 80\% & $55.58$ & $38.68$ \\
    \midrule
        \multirow{4}{*}{\rotatebox[origin=c]{90}{3d Enc}}
        & Start Relative Coord. & $59.67$ & $39.99$ \\
        & Absolute Coord. & $60.12$ & $40.98$ \\
        & BBox Coord & $60.21$ & $41.81$ \\
        & No Region Positional Enc & $58.90$ & $40.50$ \\
    \midrule
        \multirow{2}{*}{\rotatebox[origin=c]{90}{3dC}}
        & -- Context Proposals & $52.49$ & $35.64$ \\
        & -- Viewpoint Nbhd. Feat. & $58.19$ & $39.68$ \\
    \midrule
        \multirow{3}{*}{\rotatebox[origin=c]{90}{Infer}}
        & -- Viewpoint Grouping & $23.69$ & $10.17$ \\
        & -- Distance Limit & $45.38$ & $31.50$ \\
        & + Two-Step Inference & $53.80$ & $33.77$ \\
    \bottomrule
\end{tabular}
\end{table}

%% file: sections/06-conclusion.tex
We present \ourmodel, a remote grounding model that builds on a generic vision-language backbone and a ``bag of tricks'' for training and inference for this embodied task.
We set state of the art remote grounding accuracy on both REVERIE and FAO, and deploy \ourmodel\ on a physical robot.
We hope that these results serve as recommendations to the community for which tricks to apply when doing exhaustive vision-learning retrieval for remote grounding. 
Furthermore, this may help guide research for applying vision-language models to other large, geometric domains.

%% file: sections/appendix.tex
\noindent In this appendix we include:
\begin{enumerate}[noitemsep,topsep=0pt]
    \item A supplementary video demonstrating the algorithm running on a TurtleBot3 open-source platform.
    \item An ablation experiment investigating which parts of the text instructions are important for the method.
\end{enumerate}

\subsection{Home Robot Results}
We evaluate \ourmodel\ on a TurtleBot3 Burger with an attached RGBD camera running ROS nodes for coordinate frames, navigation system, and camera updates.
The robot first drives around the environment and runs Mask-RCNN on RGBD images to create regions (which have 3d coordinates from the depth).
Then, we use \ourmodel\ to select a region from among those stored, given a text query.
The 3d coordinates for the selection region are passed to the TurtleBot navigation system (ROS) which moves the robot (not a part of this work).
The user interface includes a speech-to-text system to convert a real-time microphone input into text used to select between robot operations (\texttt{go to}, \texttt{stop}, \texttt{memory\_update}, \texttt{visualize}), and passed to the \ourmodel \ system. 

The robot was able to build a map of the environment, use \ourmodel\ to identify the correct object based on a query, and use the navigation system to navigate to that target. 
Since the Mask-RCNN backbone was trained on regular images, we found that the model transferred somewhat well from the Matterport3d simulator, but we used relatively simple instructions to simplify the problem. The main challenges included: (A) making the robot stop briefly before using image or location data to ensure the data was synchronized and not blurry; (B) getting good views of objects from a 20cm tall robot platform; and (C) converting 3d object locations into navigable locations for the navigation system. 

The supplementary material includes a video of the robot doing semantic mapping, remote grounding, and navigation to a referent target given speech input by passing the coordinates from \ourmodel\ to the ROS navigation system. Note that the audio is distorted for anonymity.

\begin{table}[ht]
\setlength{\aboverulesep}{0pt}
\setlength{\belowrulesep}{0pt}
\tabcolsep=0.07cm
\centering
\small
\caption{Ablation of text information on REVERIE the val unseen split using \ourmodel.}
\vspace{-1em}
\label{tab:ablation_text}
\begin{tabular}{lroro} 
    & \multicolumn{2}{c}{\it Val Seen} & \multicolumn{2}{c}{\it Val Unseen} \\
    \bf Method & SR$\uparrow$ & RGS$\uparrow$ & SR$\uparrow$ & RGS$\uparrow$ \\ 
    \toprule
    Full Text & $\pmb{77.37}$ & $\pmb{62.97}$ & $\pmb{61.06}$ & $\pmb{43.75}$ \\
    \midrule
    No Adjectives & 75.54 & 58.05 & 58.34 & 39.70 \\
    Only Adjectives/Nouns & 75.19 & 59.38 & 58.14 & 39.28 \\
    Only Nouns & 74.07 & 52.78 & 54.79 & 34.62 \\
    No Room & 68.17 & 54.88 & 45.58 & 31.55 \\
    Only Room & 61.07 & 13.70 & 38.31 & 10.79 \\
    No Nouns & 40.97 & 23.33 & 21.27 & 9.43 \\
    \bottomrule
\end{tabular}
\end{table}

\subsection{Text Ablations}

In this section we investigate the importance of different parts of the query sentence on the final grounding performance. 
In the REVERIE dataset, queries are typically of the form ``Go to \texttt{ROOM} and \texttt{OBJECT\_DESCRIPTION}'' which allows for simple parsing on the query sentence to extract the room description (everything before the first ``and''). 
We consider the full sentence (\emph{Full Text}), removing all adjectives  (\emph{No Adjectives}), removing all except adjectives and nouns (\emph{Only Adj \& Nouns}), removing all except nouns (\emph{Only Nouns}), removing the room description (\emph{No Room}), removing all after the room description (\emph{Only Room}), and removing all nouns (\emph{No Nouns}). Each method is fully trained and tested with the given text ablation.

In Table~\ref{tab:ablation_text} we show performance on the REVERIE validation unseen set for ablations of the text. First, we note that giving only the room annotation results in relatively high performance, due to the number of provided candidate proposals visible in the correct room not being high. Second, removing everything except adjectives and nouns, results in a relatively large drop in performance, suggesting that relative locations and other prepositions are important for the final task, but not as important as the nouns.

We note that in the most extreme text ablation setting (\emph{No Nouns}) the \emph{RGS} is still $9.43\%$. This non-trivial performance may be explained by that given a limited set of candidate object regions provided by the benchmark, and a limited search distance, the search space for certain rooms may be relatively small.